\pdfoutput=1
\documentclass[11pt]{article}
\usepackage{authblk}

\usepackage[]{ACL2023}
\usepackage{booktabs}
\usepackage{caption} % DO NOT CHANGE THIS AND DO NOT ADD ANY OPTIONS TO IT
\usepackage{times}
\usepackage{latexsym}
\usepackage{algorithm}
\usepackage{amsmath}
\usepackage{algorithmic}
\usepackage{xcolor}
\usepackage{soul}
\usepackage{CJKutf8}
\usepackage{subfigure}
\usepackage{multirow}
\usepackage{microtype}
\usepackage{graphicx} % DO NOT CHANGE THIS

\usepackage[T1]{fontenc}
\usepackage[utf8]{inputenc}

\usepackage{microtype}

\usepackage{inconsolata}

\title{What Makes Pre-trained Language Models Better Zero-shot Learners?}

\author{
    Jinghui Lu\textsuperscript{\rm1},
    Dongsheng Zhu$^+$ \textsuperscript{\rm2},
    Weidong Han$^+$ \textsuperscript{\rm2},
    Rui Zhao \textsuperscript{\rm1},
    Brian Mac Namee \textsuperscript{\rm3}, 
    Fei Tan$^*$ \textsuperscript{\rm1} 
    \\
    \textsuperscript{1} SenseTime Research \\
    \textsuperscript{2} Fudan University \\
    \textsuperscript{3} School of Computer Science, University College Dublin \\
    \texttt{\{lujinghui1, zhaorui, tanfei\}@sensetime.com} \\
    \texttt{\{dszhu20, wdhan20\}@fudan.edu.cn}\\
    \texttt{\{brian.macnamee\}@ucd.ie} \\
    \texttt{}\\
    
}

\begin{document}
\begin{CJK}{UTF8}{gbsn}
\maketitle

\def\thefootnote{+}\footnotetext{Work was done during internship at SenseTime Research}

\def\thefootnote{*}\footnotetext{Corresponding author}

\begin{abstract}
Current methods for prompt learning in zero-shot scenarios widely rely on a development set with sufficient human-annotated data to select the best-performing prompt template \emph{a posteriori}. This is not ideal because in a real-world zero-shot scenario of practical relevance, no labelled data is available. Thus, we propose a simple yet effective method for screening reasonable prompt templates in zero-shot text classification: \textbf{Perple}xity Sele\textbf{ction} (Perplection). We hypothesize that language discrepancy can be used to measure the efficacy of prompt templates, and thereby develop a substantiated perplexity-based scheme allowing for forecasting the performance of prompt templates in advance. Experiments show that our method leads to improved prediction performance in a realistic zero-shot setting, eliminating the need for any labelled examples.
\end{abstract}

\section{Introduction}

\begin{table*}[!ht]
\small
\centering
\resizebox{1.0\textwidth}{!}{%
\begin{tabular}{lcccccccccc|ccc}
\toprule
\multirow{2}{*}[-3pt]{\textbf{Dataset}}  &  & \multicolumn{2}{c}{\textbf{1. {[}very/not{]} pleased.}} &  & \multicolumn{2}{c}{\textbf{2. {[}very/not{]} good.}}&  &\multicolumn{2}{c}{\textbf{3. {[}extremely/less{]} pleased.}} &  & &\multicolumn{2}{c}{\textbf{4. {[}yellow/green{]} black.}}\\[3pt] \cline{3-4} \cline{6-7} \cline{9-10} \cline{13-14} \\[-5pt]
                                  &   & \textbf{PPL}                      & \textbf{Acc.(\%)}                       &  & \textbf{PPL}                   & \textbf{Acc.(\%)}   &         & \textbf{PPL}                   & \textbf{Acc.(\%)}    &  &  & \textbf{PPL}                   & \textbf{Acc.(\%)}   \\ \midrule
\textit{DOUBAN}                                                 &  & 24.61                    & 57.12                  &  & 40.93                 & 50.98          & & 28.80  & 56.68   & & & 71.01 & 51.31    \\
\textit{WEIBO}                                                    &  & 19.78                    & 61.79                   &  & 30.37                 & 51.16         & & 22.34    & 58.35    & & & 44.45 & 50.92   \\
\textit{WAIMAI}                                                   &  & 16.44                    & 67.80                 &  & 23.34                 & 53.15        & & 19.68      & 69.72   & & & 36.07 & 48.49   \\
\textit{ECOMMERCE}                                          &  & 14.07                    & 73.12                 &  & 18.45                 & 55.68          & & 16.88   & 67.49   & & & 28.56 & 50.49   \\ \bottomrule
\end{tabular}%
}
\caption{Summary of mean perplexity scores and zero-shot accuracy of different prompt templates.}
\label{tab:pilot_study}
\end{table*}

Prompt learning has been demonstrated to be a successful remedy for challenges associated with pre-training and fine-tuning paradigm, especially in zero/few-shot scenarios~\cite{gao-etal-2021-making,schick-schutze-2021-exploiting,schick-schutze-2021-just,tam-etal-2021-improving,lu-etal-2022-rationale}. 

% The essence of prompt learning is making better use of a pre-trained language model by adding additional ``hints'' (i.e., a sequence of text)~\cite{liu2021pre} to reformulate inputs from downstream tasks in a way that mimics how examples were perceived during the pre-training phase. 

% Inspired by this, a multitude of pioneering approaches are proposed and discover that it is not just autoregressive model like GPT family can take advantage of prompt learning.

% Prompt learning originates from the practice that manipulates input texts to steer the model produce desired outputs. Researchers found out that huge language models like GPT-3~\cite{brown2020language} is too expensive to fine-tune, instead, they prefixed inputs with a few training examples (in natural language) so that the model can generalise to unseen cases, which is called in-context learning~\cite{brown2020language,xie2021explanation,liu-etal-2022-makes}.

Research has repeatedly shown that various transformer-based language models can benefit from prompt learning. For example, decoder-only models, such as those in the GPT family~\cite{brown2020language}, can better generalise to unseen cases by prefixing inputs with a few training examples (in natural language). This is known as \emph{in-context} learning~\cite{brown2020language,xie2021explanation,liu-etal-2022-makes}. Encoder-decoder models, such as T5~\cite{JMLR:v21:20-074} or BART~\cite{lewis-etal-2020-bart}, can leverage prompt learning to train versatile models for multiple tasks~\cite{khashabi-etal-2020-unifiedqa,lester-etal-2021-power}. Bidirectional encoder-only models, such as those in the BERT family~\cite{devlin2018bert,liu2019roberta}, can also manifest impressive zero-shot capacity when given proper prompts. These prompts often take the form of pre-training tasks, such as next sentence prediction~\cite{sun2021nsp} or masked language modeling (MLM)~\cite{gao-etal-2021-making,schick-schutze-2021-exploiting,schick-schutze-2021-just,tam-etal-2021-improving}---also known as \emph{cloze-style} prompt learning. 

Despite its success in encoder-only models, cloze-style prompt learning is sensitive to the specific involved templates. Multiple studies have shown that the design and choice of prompt templates greatly affect the effectiveness of zero-shot learning~\cite{tam-etal-2021-improving,pmlr-v139-zhao21c,rubin-etal-2022-learning}. Ideally, they are supposed to be as close as possible to the language used in downstream task. For example, in a sentiment analysis task, a suitable template may be \textit{``[very/not] pleased.''} that carries emotional information. However, other templates can also be used here like \textit{``[very/not] good.''}.

As shown in Table \ref{tab:pilot_study}, the performance of zero-shot learning using different sentiment-bearing templates can fluctuate significantly with different prompt templates. For the \textit{ECOMMERCE} dataset, the template \textit{``[very/not] pleased.''} achieves the best zero-shot accuracy of 73.12\%, while using the template \textit{``[very/not] good.''} results in an accuracy of only 55.68\%---which is only slightly better than random guessing. Additionally, if we choose a sentiment-irrelevant template \textit{``[yellow/green] black.''}, the accuracy significantly drops to 50.49\%, indicating that the model has no classification ability. This shows that the performance of the model is largely shaped by templates used. Therefore, selecting the most appropriate templates for downstream tasks is crucial in zero-shot learning. 

Current prompt learning methods still rely on a development set of human-annotated data for post-hoc template selection~\cite{tam-etal-2021-improving,sun2021nsp,gao-etal-2021-making,liu2021pre}: all candidate templates are evaluated using the development set and the best-performing one is chosen. This requires human annotators and does not align well with realistic zero-shot learning scenarios in which no human-annotated data is available. To address this problem, we propose a truly annotation-free perplexity-based template selection method for zero-shot prompt learning: \textbf{Perple}xity Sele\textbf{ction} (Perplection). Experiments show that Perplection is highly likely to select the most effective template accommodating true zero-shot scenarios. 

% This finding is especially encouraging because most of existing work still leverages development set to post-hoc evaluate prompt templates~\cite{gao-etal-2021-making,liu2021pre}.

In this paper, we first describe cloze-style prompt learning in Section \ref{sec:preliminaries}. Then, in Section \ref{sec:assumption}, we present our hypothesis that underpins the work and how this hypothesis can interpret some observations in the literature. Based on this hypothesis, in Section \ref{sec:method} we detail Perplection that uses perplexity to select templates \textit{a priori} without the need of any annotated examples. Section \ref{sec:pilotstudy} describes a pilot study and in Section \ref{sec:experiment}, we present realistic experiments that show that Perplection leads to performance on par with other zero-shot prompt methods that utilise a development set. 

To the best of our knowledge, we spearhead the performance screening of prompt templates for a realistic zero-shot text classification without using any human-annotated data.\footnote{Code is available at \url{https://github.com/GeorgeLuImmortal/Perplection_ACL2023}.}

% In this paper, we first describe prompt learning  in Section \ref{sec:preliminaries}. Then, in Section \ref{sec:assumption}, we present our hypothesis that the language discrepancy between prompted downstream input and pre-training corpora can impact the zero-shot performance of cloze-style prompt learning, and how this hypothesis can explain some observations in the literature. Based on this hypothesis, in Section \ref{sec:method} we propose using perplexity to select templates without the need of any annotated examples \textit{a priori}. Section \ref{sec:pilotstudy} describes a pilot study demonstrating that there is a strong correlation between perplexity and zero-shot classification performance. In Section \ref{sec:experiment}, we present more realistic experiments that demonstrate that Perplection leads to performance on par with other state-of-the-art zero-shot prompt methods, while eliminating the need for a human-annotated development set.

% \section{Related Work}

\section{Preliminaries}\label{sec:preliminaries}

In this section, we describe basic concepts and terminologies associated with prompt learning.

\subsection{Prompt Learning}\label{subsec:prompt_setup}

Note that the prompting settings and terminologies used in this work are mainly derived from the work that focuses on manual/automatic cloze-style discrete templates~\cite{gao-etal-2021-making,schick-schutze-2021-exploiting,schick-schutze-2021-just,tam-etal-2021-improving}. As text classification is well studied in prompt-based learning tasks~\cite{liu2021pre}, 
we use a simple binary sentiment analysis task to demonstrate zero-shot prompt learning in our work. 
Specifically, given an input text $x$, for example \textit{``I love this movie.''}, we are interested in classifying the sentiment polarity, $y$, of this input text, i.e., $++$ for positive or $--$ for negative. The cloze-style prompt method modifies the input $x$ and output $y$ to further exploit the capabilities of pre-trained language models. Formally, we first manipulate input text $x$ to construct a new input text, $x'$, by prefixing (or suffixing) $x$ with a \textit{template} text sequence, $t$, that includes a \textit{``[MASK]''} token. So, $x' = [x,t]$ or $x'= [t,x]$. For example, if we have an input $x = $\textit{``I love this movie.''} and we decide to prefix a template $t = $\textit{``Overall, it was a [MASK] movie.''}, $x'$ will become \textit{``Overall, it was a [MASK] movie. I love this movie.''}.

Next, $x'$ is fed into a language model to predict the likelihood with which different tokens fill \textit{``[MASK]''}. This can be achieved by applying an MLM head. Usually, researchers use prior knowledge to limit the set of potential filled tokens to those relevant to the task of interest. For example, in the sentiment classification example only two tokens would be considered: \emph{``good''} and \emph{`bad''}. We call each of these a \textit{label word}, $w$,~\cite{liu2021pre}. Finally, we define a mapping function (or \textit{verbaliser})~\cite{liu2021pre}, $v$, to reverse the predicted label word back to the target $y$, for example  \{\textit{good}:$++$, \textit{bad}:$--$\}. In this way the prompting method unifies a binary classification objective into an MLM objective, reusing a MLM head to perform zero-shot prediction. 

% Table \ref{tab:notations} summarises the terminologies, notations, examples and descriptions of above prompting procedure.

\subsection{Language Discrepancy and Objective Gap}\label{subsec:assumption1}

Previous research~\cite{liu2021pre} has shown that prompt learning can help pre-trained language models better adapt to downstream tasks by closing the gap between pre-training and the downstream task. To be specific, prompt learning allows pre-trained language models to take on a greater role in prediction, rather than just extracting features. 

In light of the above finding, we identify two obstacles to combining pre-training and a downstream task: \textit{language discrepancy} and the \textit{objective gap}. The objective gap describes the difference in training objectives between pre-training (e.g., next sentence prediction or MLM) and a downstream task (e.g., sequence classification or sequence labelling). Language discrepancy refers to the linguistic differences between a pre-training corpus and downstream datasets, including different vocabularies, word frequencies, syntactic arrangements, etc.

% \begin{table*}[!t]
% \centering
% \resizebox{.9\textwidth}{!}{%
% \begin{tabular}{lcll}
% \toprule
% \textbf{Name} & \textbf{Notation} & \textbf{Example}                   & \textbf{Description}         \\ \midrule
% \textit{Input}                    & $x$                                     & I love this movie.                                     & Input text sequence                              \\
% \textit{Output}                   & $y$                                     & $++$ (positive), $--$ (negative)                          & Output target                                    \\
% \textit{Template}                 & $t$                                     & Overall, it was a {[}MASK{]} movie.                    & Template text sequence that contains {[}MASK{]}  \\
% \textit{Prompted input}           & $x'$                                     & Overall, it was a {[}MASK{]} movie. I love this movie. & Input text sequence after prompting              \\
% \textit{Label word}               & $w$                                     & good, bad                                              & Word that fills {[}MASK{]}                       \\
% \textit{Verbaliser}               & $v$                                     & \{good:$++$, bad:$--$\}                                    & Function that reverses label word back to target \\ \bottomrule
% \end{tabular}%
% }
% \caption{Summary of terminologies, notations, examples and descriptions, adapted from~\citet{liu2021pre}.}
% \label{tab:notations}
% \end{table*}

\section{Hypotheses}\label{sec:assumption}

This section proposes two hypotheses that underpin our work, and describes the way they interpret observations in the literature.

\subsection{Hypothesis I: Cloze-style Prompting Offers a Better Feature Space} \label{subsect: hypothesis1}

Our first hypothesis is that the use of a cloze-style prompt in text classification alters the input data distribution in a way that encourages the input data to be more effectively represented in a new feature space. To illustrate this, Figure \ref{fig:umap} presents a UMAP~\cite{McInnes2018} visualisation of a sentiment analysis dataset, \textit{WEIBO}, with and without prompt templates. It is obvious that after being prompted with a task-specific template, \textit{``[very/not] pleased.''}, data from different classes is much better separated within the resultant feature space (Figure \ref{subfig:w_template}) than when no prompt template is used (Figure \ref{subfig:wo_template}). This shows that a pre-trained language model can inherit zero-shot capabilities when given appropriate prompts, even without using any human-annotated examples. 

So how do pre-trained language models construct such effective feature spaces? We conjecture that this is because some knowledge of downstream tasks has been implicitly encoded into models through pre-training (e.g., MLM for encoder-only model or Next Word Prediction for decoder-only models). Prompt learning finds a method to uncover the knowledge obtained in pre-training. Therefore, in this paper, we refer to this feature space as the \textit{``pre-trained feature space.''}

\subsection{Hypothesis II: Language Discrepancy Measures the Efficacy of Prompting}\label{subsec:language discrepancy}

Additionally, we aim to understand what makes a template effective at forming a useful pre-trained feature space. We believe that the difference in language between pre-training corpora and downstream datasets after prompting can be used to assess the effectiveness of templates.

\begin{figure*}[!t]
    \centering
    \subfigure[No template]{\includegraphics[width=.29\textwidth]{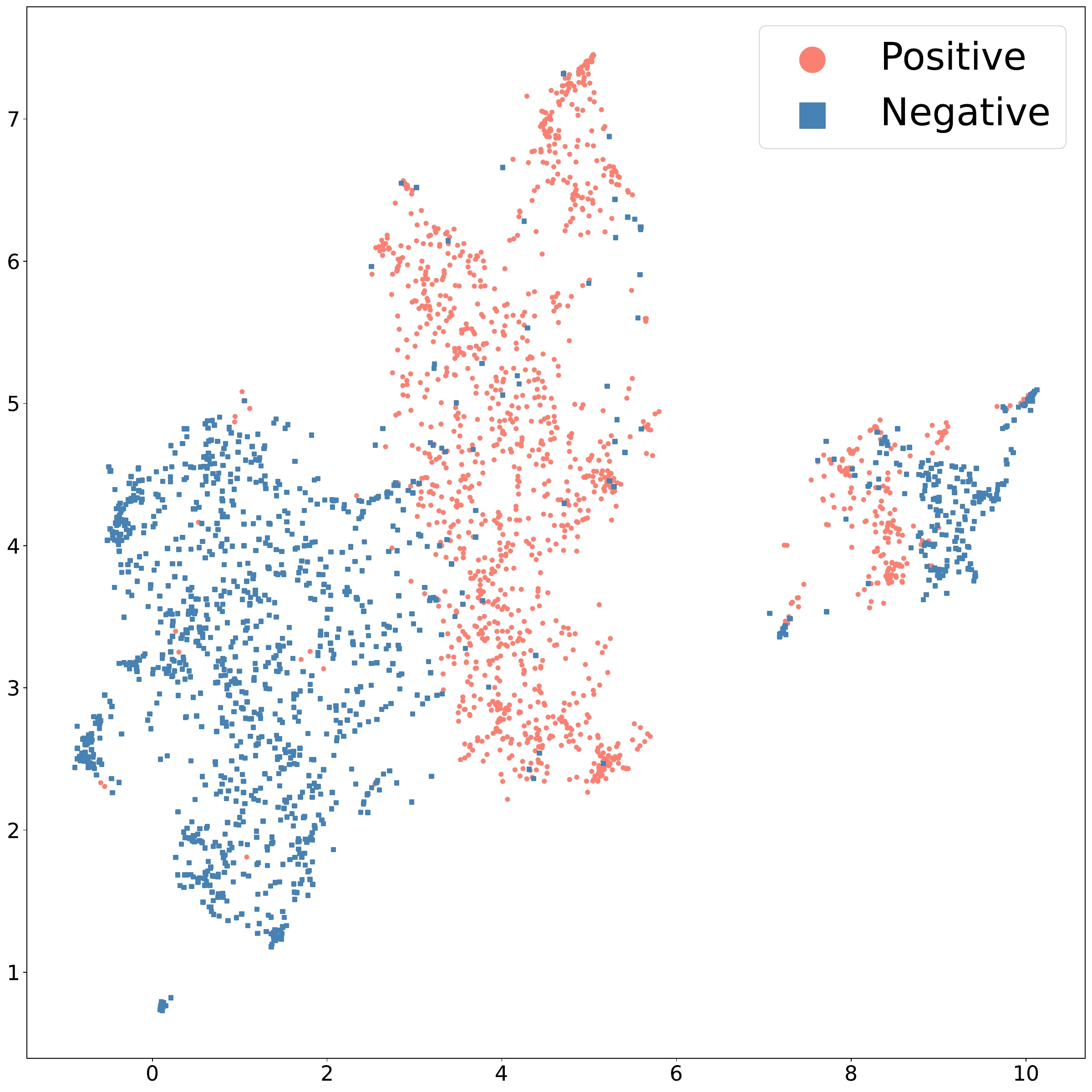}\label{subfig:wo_template}}
    \subfigure[\texttt{[very/not]} pleased.]{\includegraphics[width=.29\textwidth]{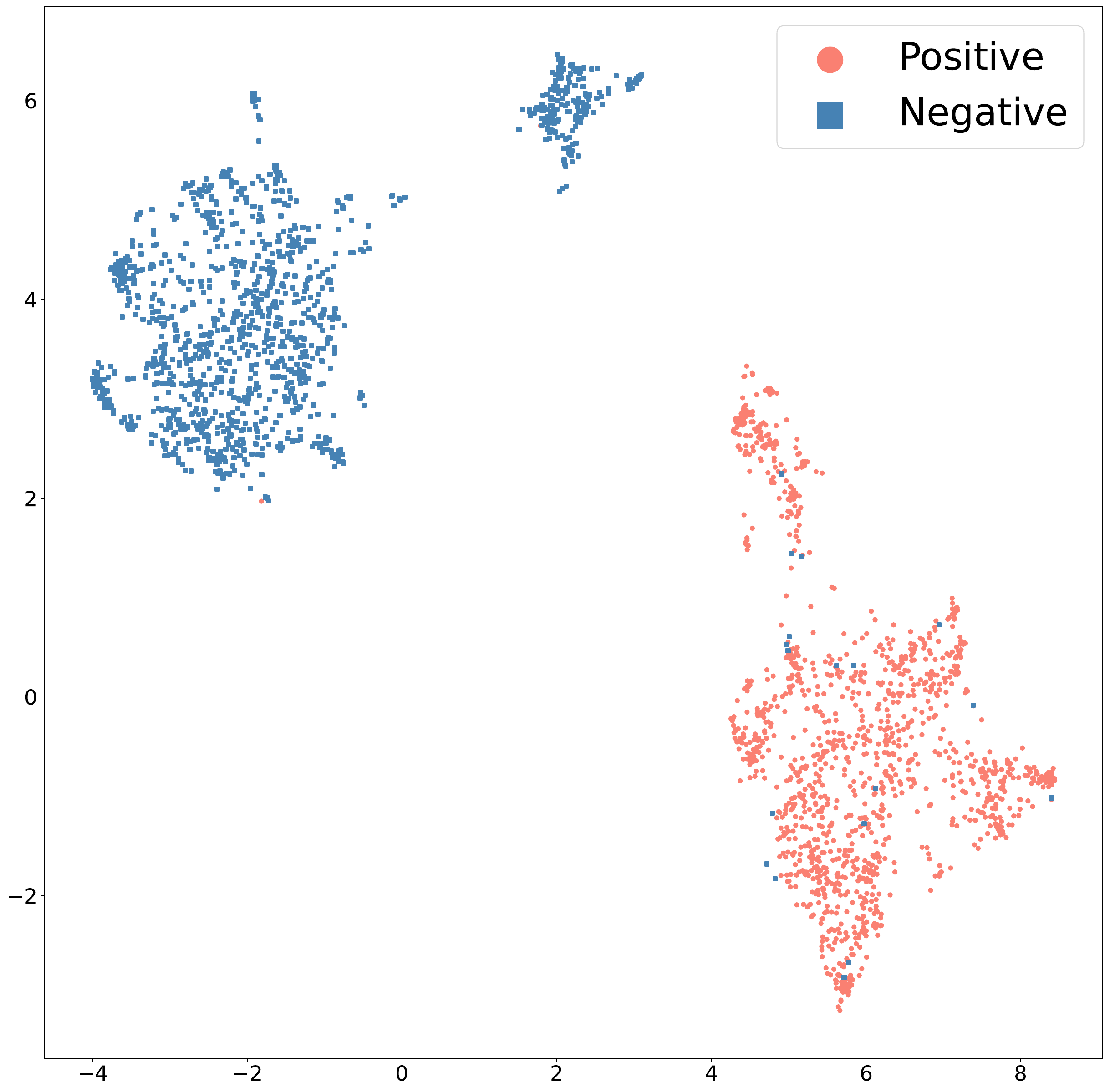}\label{subfig:w_template}}
    \subfigure[\texttt{[yellow/green]} black.]{\includegraphics[width=.29\textwidth]{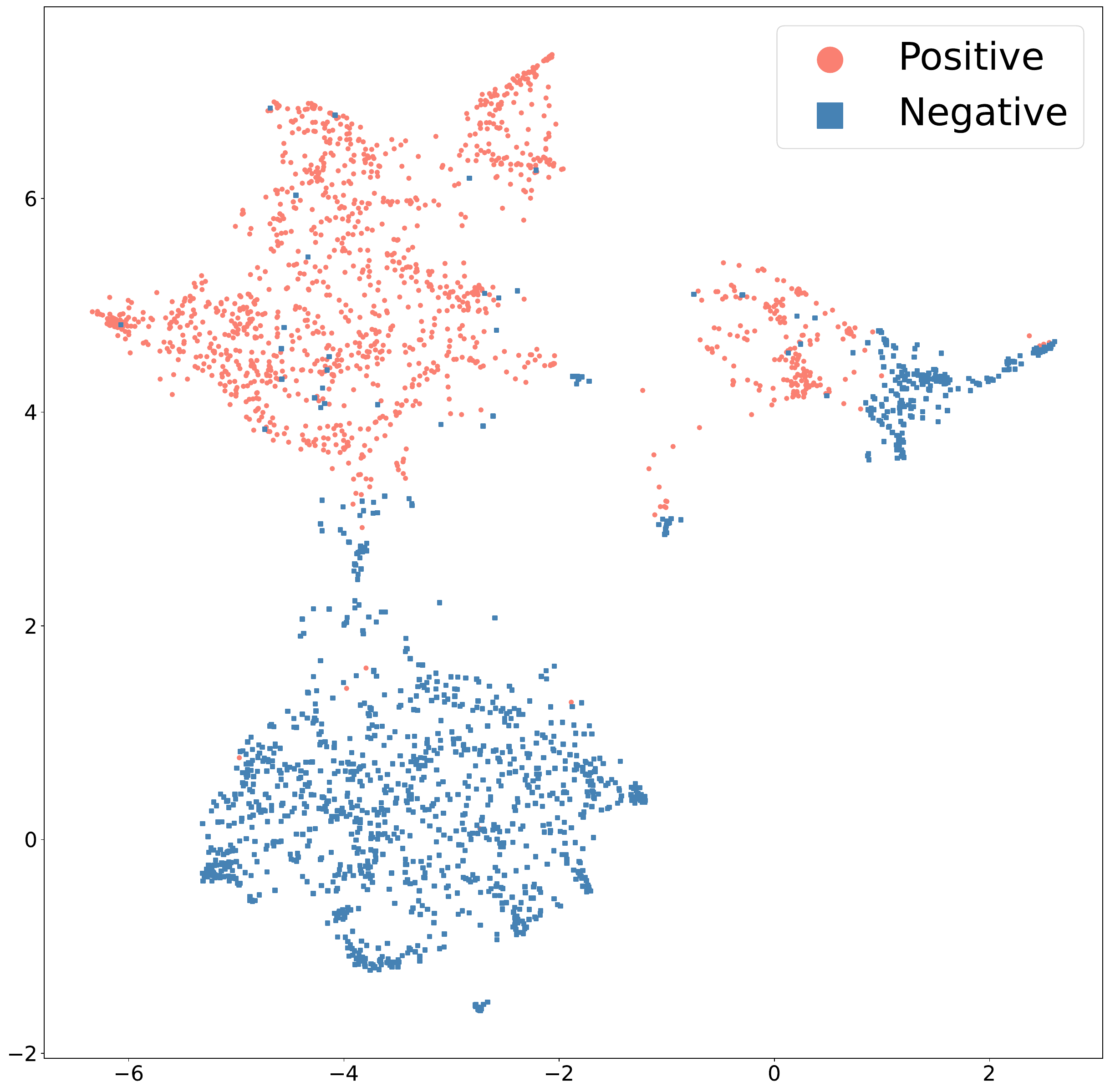}\label{subfig:bad_template}}
    \caption{UMAP visualisation of a sentiment analysis dataset \textit{WEIBO}: (a) no template, (b) task-relevant template, and (c) irrelevant template. (Best viewed in color.)}
    \label{fig:umap}
\end{figure*}

Figure \ref{subfig:bad_template} shows an example. When the text inputs are given a prompt that is unlikely to be used in sentiment analysis texts, \textit{``[yellow/green] black.''}, the data from different classes is not well separated in the feature space (as compared to Figure \ref{subfig:w_template}). We believe that this is because models rarely encounter the text \emph{``yellow black''} or \emph{``green black''} prefixed in a sentiment-bearing text in the pre-training corpora, and that this language discrepancy limits the model's ability to effectively represent the data. In contrast, expressions like \textit{``[very/not] pleased.''} (Figure \ref{subfig:w_template}) are often used in context related to emotions and therefore appear more frequently together with sentiment-bearing text in the pre-training corpora. This makes it easier for the model to form a useful pre-trained feature space.\looseness=-1

Broadly speaking, we suppose that the objective gap has been greatly reduced by reformulating the downstream task to use a prompt in text classification. The inconsistency is largely due to the language differences between the pre-training data and the downstream data. Using prompt templates helps to align the downstream text with the text in a pre-training corpus with respect to language discrepancy. The smaller the language discrepancy between the pre-training data and the downstream data that are being prompted, the more likely it is that the data will be represented well in the feature space, resulting in better zero-shot performance.\looseness=-1

% \subsection{Discussion}\label{subsec:heuristic}

% Our hypotheses partially suggest explanations for several observations in the literature. For example, several studies have shown evidence that using manually crafted domain-related templates~\cite{radford2021learning} or domain-specific tokens as the initial prompt in soft prompting~\cite{lester-etal-2021-power,zhou2022learning,khattak2022maple} can lead to better classification performance. The addition of domain-specific tokens is more likely to align downstream data into  that the pre-trained feature space, leading to a better initialisation for further optimisation. Essentially, sentiment analysis is a simple form of multi-class classification where sentiment polarity (i.e., positive or negative) is the domain of interest. As such, researchers usually use prompt with sentiment label words such as \textit{``It was a [good/bad] movie.''} in experiments~\cite{liu2021pre}.

\begin{figure}[t]
\centering
\includegraphics[width=.49\textwidth]{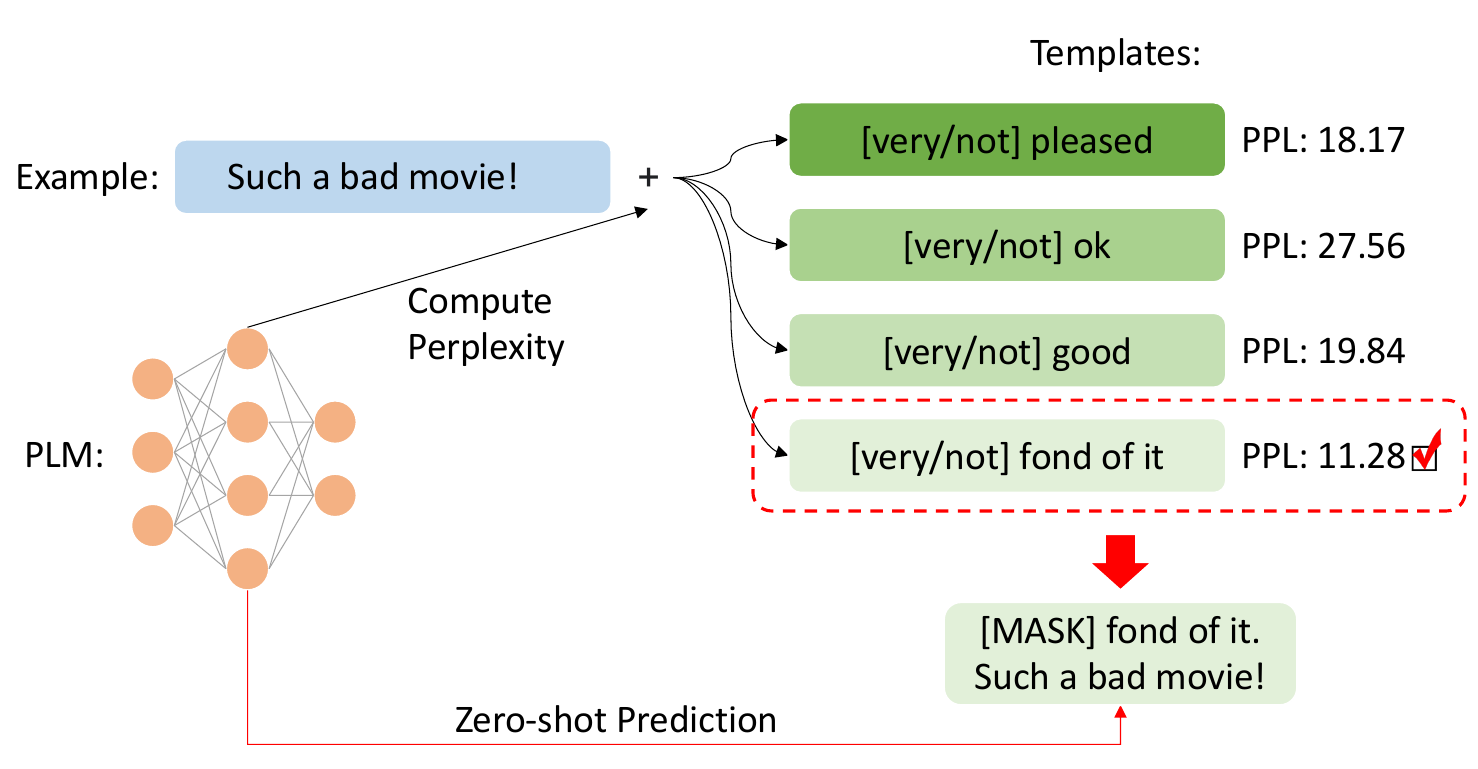}
 \caption{The procedure of the Perplection approach.}\label{fig:overview}
\end{figure}

\section{Method} \label{sec:method}

% \subsection{Extensions}

As discussed in Section \ref{sec:assumption}, a heuristic approach can be employed to select the most effective templates in zero-shot text classification. One way to do this is to utilise language discrepancy to ``forecast'' the performance of different prompt templates. Specifically, the prompt template that results in the lowest language discrepancy when prefixed to a given input text can be considered the most effective. However, how can the language discrepancy between downstream text and pre-training corpora be measured? In this study, we propose using perplexity~\cite{brown-etal-1992-estimate} as an approximation of language discrepancy.

Perplexity is one of the most common metrics for evaluating language models, and is defined as the exponential average negative log-likelihood of a sequence:

\begin{equation}
\small
\operatorname{PPL}(x)=\exp \left\{-\frac{1}{t} \sum_{i}^{t} \log p_{\theta}\left(x_{i} \mid x_{<i}\right)\right\}
\label{eq:ppl}\end{equation}
    
\noindent where $x = [x_{1}, x_{2}, ..., x_{t}]$ is a tokenised text sequence; and $\log p_{\theta}\left(x_{i} \mid x<i\right)$ is the log-likelihood of the $i^{th}$ token conditioned on the preceding tokens 
$x<i$ computed by a language model. Intuitively, given a certain language model, lower perplexity for a corpus of sentences indicates a model is familiar with that corpus. Basically, the language model with the lowest perplexity is chosen as the most reliable proxy for modelling the distribution of the pre-training corpus. 

Analogously, we assume that prompt templates resulting in low perplexity when prefixed to a given input are likely to be effective templates, eliminating the need for a human-annotated development set, which is required in most previous work~\cite{liu2021pre,lester-etal-2021-power,gao-etal-2021-making}. Specifically, as shown in Figure \ref{fig:overview}, we prefix original input $x$ with various prompt templates to form new prompted texts. For each template, since we have two label words (i.e., \emph{``very''} and \emph{``not''}), one original input $x$ will generate two prompted texts (i.e., \emph{``Very pleased. Such a bad movie!''} and \emph{``Not pleased. Such a bad movie!''}). Then we compute the mean perplexity score of these two prompted texts as the score for the template. Finally, the template (where the label words will be replaced with \emph{"[MASK]"} token) with lowest score is selected to be prefixed to the original input, constructing new input $x'$ (i.e., \emph{``[MASK] pleased. Such a bad movie!''}) to perform a zero-shot prediction. This is quite different from previous methods with dataset-specific~\cite{gao-etal-2021-making,sun2021nsp} or class-specific templates~\cite{zhou2022learning}. We refer to the method as \textbf{Perple}xity Sele\textbf{ction} (Perplection).

% Previous work~\cite{liu2021pre,lester-etal-2021-power,gao-etal-2021-making} relies on development set to post-hoc choose the best performing prompt templates. 

\section{Pilot Study} \label{sec:pilotstudy}

The aim of the pilot study described in this section was to qualitatively validate the hypotheses proposed in Section \ref{sec:assumption}, and to examine the utility of perplexity as a metric for screening prompt templates (another study that examines the utility of perplexity is presented in Appendix \ref{appendix:reverse_exp}). To this end, we manually curated four prompt templates as shown in Table \ref{tab:pilot_study}. We then analysed the perplexity and zero-shot performance of each template, seeking to determine whether there is a correlation between perplexity and zero-shot performance. \looseness=-1

\subsection{Datasets} \label{subsec:pilot_study:datasets}

We conducted the pilot study using four publicly available Chinese sentiment analysis datasets from various domains. These datasets are: \textit{DOUBAN}, a movie review dataset; \textit{WEIBO}, a social media comment dataset; \textit{WAIMAI}, a takeaway comment dataset; \textit{ECOMMERCE}, an e-commerce dataset.

% Datasets are in Chinese and we have  translated the examples shown in this paper to facilitate understanding.\looseness=-1

% Each dataset is split into five training sets/one development set/one test set.

\subsection{Perplexity}

We use the Chinese RoBERTa model\footnote{Available at \url{https://huggingface.co/hfl/chinese-roberta-wwm-ext.}} as the backbone pre-trained model. Given a pre-trained language model, we use it to compute the mean perplexity of downstream datasets that are being prompted, to approximate the language discrepancy. That is, lower perplexity indicates smaller language discrepancy between the pre-training corpus and the prompted downstream dataset.

Note that perplexity, as originally defined, applies specifically to causal language models (i.e., autoregressive language models). As suggested in previous work~\cite{liu2019roberta,salazar-etal-2020-masked}, perplexity for bidirectional models like BERT/RoBERTa can be made analogous to that for causal language models by replacing $\log p_{\theta}\left(x_{i} \mid x<i\right)$ with $\log p_{\theta}\left(x_{i} \mid c\right)$ in Equation \ref{eq:ppl}. Here, $c$ refers to the context text, which is the whole sentence except for the $i^{th}$ token. This suggests that the perplexity of each token is not only conditioned on the preceding tokens but also the succeeding tokens. We added a template to each example, replaced the \textit{``[MASK]''} with label words from the prediction problem, and calculated the average perplexity for each example. We then averaged the perplexity scores of all examples to get the overall perplexity of the dataset.

During preliminary experiments, however, we found that this definition of perplexity has the drawback of favouring longer sentences. That is, a sentence is assigned a lower perplexity, not because the pre-trained language model is more able to model this sentence (i.e., low language discrepancy), but rather because the text is longer. We conjecture that this is due to the penalty term in Equation \ref{eq:ppl} that divides the sum of log-likelihood by the sequence length $t$. The detail of our preliminary experiments regarding perplexity are provided in Appendix \ref{appendix:perplexity}. The focus of this pilot study, however, is to illustrate the impact of language discrepancy rather than finding useful measures of perplexity. So, to mitigate against the drawbacks of the perplexity definition the four datasets used in our experiments were subsampled to include only sentences with  between 14 and 15 words, as well as to enforce a 50:50 class balance. Also, all hand-crafted templates have similar lengths (in Chinese).

\subsection{Zero-shot Result Analysis} 

The accuracies achieved using different prompt templates for four datasets are shown in Table \ref{tab:pilot_study}. These results demonstrate that prompt learning can equip a pre-trained language model with zero-shot capability when proper templates are provided. However, the performance of Template 4 (i.e., \textit{``[yellow/green] black''}) demonstrates that ``unusual'' prompting (i.e., texts that models are unlikely to see during pre-training) has limited contribution to zero-shot prediction, which is consistent with our expectation.

To conclude, the results of the pilot study verify our hypothesis that in prompt learning, task-related templates are more useful in shaping a good pre-trained feature space. The big difference between zero-shot performance across different prompting approaches in the pilot study shows that it is crucial to search for ideal prompt templates in prompt learning. We argue that this problem can be addressed by using perplexity as discussed in the following subsection.

\subsubsection{Perplexity Analysis} 

% Though these experiments do not mainly aim for reflecting the relationship between language discrepancy and model performance, 

Table \ref{tab:pilot_study} also conveys a very clear message that as perplexity goes up, the zero-shot performance becomes worse. For example, the perplexity of Template 1 decreases from 24.61 (\emph{DOUBAN}), to 19.78 (\emph{WEIBO}), to 16.44 (\emph{WAIMAI}), to 13.71 (\emph{ECOMMERCE}); while the zero-shot accuracy consistently increases from 57.12 (\emph{DOUBAN}), to 61.79 (\emph{WEIBO}), to 67.80 (\emph{WAIMAI}), to  73.12 (\emph{ECOMMERCE}). This pattern can also be observed for Templates 2 and 3. Furthermore, when comparing sentiment-bearing templates (Templates 1-3) to the sentiment-irrelevant template (Template 4) across datasets, it is evident that the sentiment-irrelevant template consistently yields the highest perplexity and the lowest accuracy. The experimental results can partially verify our hypotheses that as the language discrepancy decreases (i.e., lower perplexity), it is easier for prompts to align downstream data to a pre-trained feature space. The next section describes experiments that show how the Perplection approach takes advantage of this.

% Interestingly, if we compare across two templates (i.e., the third and the fourth column) for the same dataset, we can find out that the template with lower perplexity implies a higher accuracy. For example in DOUBAN, perplexity is 24.25 vs 39.70 while the accuracy is 65.13 vs 50.98. These observations is consistent in all datasets. However, as discussed in Section \ref{subsec:assumption2}, the zero-shot prediction performance is depends on the synergy between templates, label words and pre-trained language models. We can presume that a template with low perplexity is more likely to form a better pre-training distribution, facilitating the choice of label words. We leave this exploration for future work.

\section{Experiments} \label{sec:experiment}

\begin{table*}[!ht]
\small
\centering
\resizebox{1.0\textwidth}{!}{%
\begin{tabular}{llllll|lll}
\toprule
 \multicolumn{1}{c}{}& \multicolumn{5}{c}{\textbf{Binary Classification}} & \multicolumn{3}{c}{\textbf{Multi-class Classification}} \\\midrule
\textbf{Manual Templates} & \textbf{DOUBAN} & \textbf{WEIBO} & \textbf{WAIMAI} & \textbf{ECOMMERCE} & \textbf{EPRSTMT} & \textbf{TNEWS} & \textbf{CSLDCP} & \textbf{IFLYTEK} \\ \midrule
% \textit{Zero-shotG}~\cite{xu2021fewclue}               & 49.86               & 49.94              & 49.73               & 49.91                  &57.50 & 37.00   \\
% \textit{Zero-PET}~\cite{xu2021fewclue}               &                &               &                &                  & &    \\
% \textit{NSP-BERT}~\cite{sun2021nsp}                 & 59.86           & \textbf{72.69} & \textbf{76.71}  & 82.63       & 86.90 & 51.90       \\ \midrule

% \textbf{Manual Templates} &  &  &  &  \\ \midrule
\textit{MRandomB}            & 57.89           & 60.37         & 69.31           & 71.61    & 62.26 &    24.90  & \textbf{27.57} & 45.29   \\
\textit{MPerplectionB}   &    \textbf{59.86}      &    \textbf{64.71}      &    \textbf{\underline{79.01}}      &      \textbf{81.78}         & \textbf{67.86} &   \textbf{29.05}    & 23.36 &  \textbf{47.76} \\\midrule
\textit{MRandomR}            & 55.72           & 60.47          & 66.43           & 72.49    & 67.40 &   24.56    &  26.95 & 44.94 \\
\textit{MPerplectionR}               & \textbf{\underline{60.74}}  & \textbf{66.50}          & \textbf{75.49}           & \textbf{\underline{85.12}} & \textbf{\underline{76.89}} & \textbf{\underline{35.92}}  & \textbf{36.75} & \textbf{\underline{55.88}}  \\ \midrule
\textbf{Automatic Templates}                       &                 &                &                 &                    \\ \midrule
\textit{ARandomB}              & \textbf{54.27}          & 52.39          & \textbf{56.57}          & 58.52     & 53.18 & \textbf{28.45}    & 37.77&  51.17  \\
\textit{APerplectionB}                 & 53.07           & \textbf{57.60}          & 53.15          & \textbf{68.16}   & \textbf{55.24}  & 25.67  &\textbf{38.74}&  \textbf{51.29}     \\\midrule
\textit{ARandomR}              & 53.83           & 52.50          & 56.02           & 58.83     & \textbf{53.14} & 25.72   & \textbf{\underline{41.31}} &  49.29    \\
\textit{APerplectionR}                 & \textbf{59.21}           & \textbf{\underline{67.04}}          & \textbf{72.19}           & \textbf{73.94}   & 53.11 &   \textbf{27.34}  & 39.31 &  \textbf{51.18}  \\\bottomrule

\end{tabular}%
}
\caption{Results for text classification datasets. \textit{B} and \textit{R} stand for BERT and RoBERTa models, respectively. The \textbf{bolded} entries represent the superior performance of the Perplection variant compared to its random counterpart. The \underline{underlined} entries denote the top-performing method among all variants.}
\label{tab:method}
\end{table*}

\begin{table*}[!ht]
\small
\centering
\resizebox{1.0\textwidth}{!}{%
\begin{tabular}{llllll|lll}
\toprule
 \multicolumn{1}{c}{}& \multicolumn{5}{c}{\textbf{Binary Classification}} & \multicolumn{3}{c}{\textbf{Multi-class Classification}} \\\midrule
\textbf{State-of-the-art Methods} & \textbf{DOUBAN} & \textbf{WEIBO} & \textbf{WAIMAI} & \textbf{ECOMMERCE} & \textbf{EPRSTMT} & \textbf{TNEWS} & \textbf{CSLDCP} & \textbf{IFLYTEK} \\ \midrule
% \textit{Zero-shotG}~\cite{xu2021fewclue}               & 49.86               & 49.94              & 49.73               & 49.91                  &57.50 & 37.00   \\
\textit{Zero-PET}~\cite{schick-schutze-2021-exploiting}               &      51.64          &      51.52         &       56.71         &        60.82          & 59.51& 22.58  & 32.19 &75.29 \\
\textit{NSP-BERT}~\cite{sun2021nsp}                 & \textbf{60.85}           & \textbf{68.58} & \textbf{83.69}  & \textbf{91.11}       & \textbf{79.67} & \textbf{49.55}  & \textbf{48.43}    & \textbf{78.82} \\ \midrule
\textit{MPerplectionR}               & 60.74  & 66.50          & 75.49           & 85.12 & 76.89 & 35.92  & 36.75 &55.88  \\ \bottomrule

\end{tabular}%
}
\caption{A comparison of the performance of Perplection with that of recent state-of-the-art methods.}
\label{tab:vs_sota}
\end{table*}

\begin{table*}[!t]
\centering
\small
\begin{tabular}{llll}
\toprule
\textbf{ID} &  \textbf{Manual Template (binary)} &\textbf{Manual Template (multi-class)} & \textbf{Automatic Template (\emph{TNEWS})}                          \\ \midrule
1                  &       {[}MASK{]} satisfied                            & This belongs to {[}MASK{]}                                       &              New {[}MASK{]}\textcolor{red}{：}             \\
2         & {[}MASK{]} fond of it        & The words belong to {[}MASK{]}                                                            &     Good {[}MASK{]}\textcolor{red}{：}                               \\
3  &  {[}MASK{]} pleased             & Actually it is {[}MASK{]}                                                         & \textcolor{red}{《}{[}MASK{]}\textcolor{red}{》}  \\
4           &   {[}MASK{]} pretty good &      Probably it is {[}MASK{]}                             &     Good {[}MASK{]}\textcolor{red}{！}        \\
5              &          {[}MASK{]} happy                           &    The direction is {[}MASK{]}                                          &    Net {[}MASK{]}\textcolor{red}{：}                 \\
6              &           {[}MASK{]} good                          &  This is due to    {[}MASK{]}                                          &       Good {[}MASK{]}|            \\
7              &       {[}MASK{]} ok                              &     Put it into {[}MASK{]}                                          &        New {[}MASK{]}|           
\\
8              &       -                             &       It means  {[}MASK{]}                                      &       . {[}MASK{]}\textcolor{red}{！}            
\\
9              &       -                             &       Obviously counted as {[}MASK{]}                                        &        Good {[}MASK{]}\textcolor{red}{，}            
\\
10              &       -                             &       Obviously it is {[}MASK{]}                                        &      In {[}MASK{]}\textcolor{red}{，}              
\\
11              &       -                             &       -                                    &      New {[}MASK{]}:              
\\
\bottomrule
\end{tabular}%

\caption{The templates used for binary sentiment analysis and topic multi-class classification datasets. Due to space considerations, for automatically generated templates, we only present templates used in \emph{TNEWS}. The \textcolor{red}{red text} denotes Chinese punctuation marks. More details are provided in Appendix \ref{appendix:templates}.}
\label{tab:template}
\end{table*}

In this section, we demonstrate the proposed Perplection approach in a more realistic and useful experimental setting to verify \textit{whether we can use language discrepancy to forecast the efficacy of prompt templates for zero-shot classification}. 

% We do not investigate a few-shot setting since the previous experiments have demonstrated the impact of templates is much reduced when model is optimised by few labelled data.
\subsection{Datasets}

In addition to the datasets mentioned in Section \ref{subsec:pilot_study:datasets}, we also utilise four text classification datasets from the \textit{FewCLUE} benchmark~\cite{xu2021fewclue}: \textit{EPRSTMT} (e-commerce comment sentiment analysis), \textit{CSLDCP} (scientific literature subject classification), \textit{TNEWS} (news classification), and \textit{IFLYTEK} (APP description topic classification). To evaluate whether Perplection can be extended to other languages, we also evaluate Perplection on three English datasets: \emph{SST-2} (sentiment analysis)~\cite{wang-etal-2018-glue}, \emph{TweetEval} (hate speech detection)~\cite{barbieri-etal-2020-tweeteval}, and \emph{AG News} (multi-class topic classification)~\cite{Zhang2015CharacterlevelCN}. Note that in contrast to the pilot study, in these experiments we did not subsample the datasets to make their sentences the same length. \looseness=-1

% Of these datasets, \textit{CSLDCP}, \textit{TNEWS}, and \textit{IFLYTEK} are topic-based multi-class classification datasets, providing a comprehensive text classification benchmark for our study.

\subsection{Setup}

% For the sentiment analysis datasets used in our experiments, we fix the verbaliser to be \{\textit{very}: $++$, \textit{not}: $--$\} and manually generate seven sentiment-bearing templates. For the multi-class classification datasets, we manually generate ten templates and all class names (will be used as label words) are converted into the same length (see Appendix \ref{appendix:implemenation}). All manually crafted templates are presented in Table \ref{tab:template}. Manual templates for English datasets can be seen in Appendix \ref{appendix:implemenation}. 
All manually crafted templates are presented in Table \ref{tab:template}. All the verbalisers and manual templates for English datasets can be seen in Appendix \ref{appendix:implemenation}. 
We perform Perplection based on these manually designed templates (\textbf{MPerplection}). If perplexity is an ideal metric, the performance of this method will be better than random template-example matching (\textbf{MRandom}). We then construct a more aggressive setting where templates are generated automatically by LM-BFF algorithm~\cite{gao-etal-2021-making} (more detail is included in Appendix \ref{appendix:templates}) and apply similar template selection procedures to those described for manually crafted templates. These are dubbed \textbf{APerplection} and \textbf{ARandom}. In order to obtain a robust assessment of the random variants, we conduct five independent runs of the experiments using different random seeds and report the average results. Note that both manually crafted and automatically generated templates are constructed to have similar lengths.

% The automatic template generation method is derived from~\citet{gao-etal-2021-making} which results in between 11 and 73 templates for each dataset after removing duplicates (more detail is included in Appendix \ref{appendix:templates}). Note that both manually crafted and automatically generated templates are constructed to have the similar lengths in their original languages.
% During inference, we perform instance-level template-example matching. In other words, for each example, we select one template from the pool with the lowest perplexity score and use it to make a prediction (\textbf{MPerplection}). This is quite different from previous methods with dataset-specific or class-specific templates. If perplexity is an ideal metric, the performance of this method will be better than random template-example matching (\textbf{MRandom}). \looseness=-1

% Using automatically generated templates, we 

We report the results based on both RoBERTa and BERT\footnote{\url{https://huggingface.co/bert-base-chinese}} to demonstrate the proposed method is agnostic to the pre-trained model used. We also report the performance of another two state-of-the-art zero-shot prompting-based methods: \textbf{NSP-BERT}~\cite{sun2021nsp}, and \textbf{Zero-PET}~\cite{schick-schutze-2021-exploiting,xu2021fewclue}. They are strong baselines whose settings comply with the corresponding work (further implementation details are provided in Appendix \ref{appendix:implemenation}).

\subsection{Results} \label{subsect:exp_results}

\paragraph{Comparison to random baselines:} The results of the Perplection variants and their corresponding random counterparts were compared in Table \ref{tab:method}. It can be seen that when using manually crafted templates with both BERT and RoBERTa, Perplection was able to actively select more useful templates compared to the random selection, as indicated by the significant improvement in performance (MRandomB vs. MPerplectionB and MRandomR vs. MPerplectionR). Also, when using automatically generated templates, Perplection is able to choose more effective templates, particularly when using RoBERTa (ARandomR vs. APerplectionR). These findings suggest that the templates selected by perplexity are more useful and deliver better performance. However, results also show that Perplection is less effective when automatically generated templates are used, which will be discussed in the next section.\looseness=-1

% First of all, both Auto-PPL and Manual-PPL outperform their random sampling counterparts (i.e., Auto-Random and Manual-Random, respectively) by a clear margin, which demonstrates that the prompting settings selected by perplexity are more useful in mapping downstream data into pre-trained feature space, thus, reaching better performance.\looseness=-1 
\paragraph{Manual templates vs. automatic templates:} Table \ref{tab:method} shows that variants using manually generated templates outperform their counterparts using automatically generated templates. We conjecture that the poor quality of automatically generated templates may hinder the performance of Perplection. In other words, the pool of automatically generated templates may be insufficient in diversity for Perplection to have an impact. 

As illustrated in Table \ref{tab:template}, the majority of automatic template texts display minimal variations and lack coherence, which is in stark contrast to the manual templates. In this case, templates tend to generate similar perplexities, leading to little distinction between them based on perplexity. To illustrate this, we report the standard deviation of perplexity for both manual templates and automatic templates in Table \ref{tab:ppl_std}. It can be observed that for all datasets, the standard deviation of perplexity for manual templates is higher than that of automatic templates, showing that perplexity is more useful when the templates are of higher diversity.

\begin{table}[!t]
\small
\centering
\resizebox{.5\textwidth}{!}{%
\begin{tabular}{lllll}
\toprule
\textbf{Datasets} &  \textbf{EPRSTMT}  & \textbf{TNEWS}     & \textbf{CSLDCP} & \textbf{IFLYTEK}                      \\\midrule 
Manual Std. & \textbf{57.26} & \textbf{68.39} & \textbf{1.51} & \textbf{6.28} \\
Automatic Std. & 32.78 & 50.50 & 1.45 & 5.46 \\
\bottomrule

\end{tabular}%
}
\caption{Comparison of perplexity standard deviation.}
\label{tab:ppl_std}
\end{table}

\begin{table}[!t]
\small
\centering
\resizebox{.5\textwidth}{!}{%
\begin{tabular}{lllll}
\toprule
\textbf{Datasets} &  \textbf{SST-2}  & \textbf{TweetEval}     & \textbf{AG News} & \textbf{Avg.}                      \\\midrule 
MRandomB & 67.13 & 52.39 &  41.31 & 53.61  \\
MPerplectionB & \textbf{\underline{68.17}} & \textbf{53.67} & \textbf{\underline{43.92}} & \textbf{\underline{55.25}} \\\midrule
MRandomR & \textbf{58.79} & 54.65 & 36.85 & 50.09 \\
MPerplectionR & 57.96 & \textbf{\underline{55.16}} & \textbf{42.30} & \textbf{51.81} \\
\bottomrule

\end{tabular}%
}
\caption{Results for three English classification datasets.}
\label{tab:eng_result}
\end{table}

It is suspected that the quality of the automatically generated templates is constrained by the capacity of the pre-trained T5 model. We believe that this can be improved by changing the T5 backbone or resorting to other methods that automatically generate templates using annotation information \cite{lester-etal-2021-power,liu2021gpt,li-liang-2021-prefix,liu-etal-2022-p}. We leave these explorations for future work.

\paragraph{Comparison to state-of-the-art approaches:} We compare our best performing method (MPerplectionR) with other state-of-the-art zero-shot methods, results are shown in Table \ref{tab:vs_sota}. We find that the performance of Perplection consistently surpasses Zero-PET for all datasets by a large margin except for \emph{TNEWS}, and is competitive with NSP-BERT in some datasets such as \emph{DOUBAN} (60.74 vs. 60.85). Note that both Zero-PET and NSP-BERT used a human-annotated development set to select the most suitable templates while Perplection does not require any annotated data. 

For the \emph{IFLYTEK} dataset, Perplection seems less competitive as compared to Zero-PET and NSP-BERT. Specifically, the latter two methods heavily rely on the post-hoc selected template \textit{``This is a [MASK] app.''} (see Appendix \ref{appendix:implemenation}) with the development set quite close to target domain of interest, whereas Perplection has more generic templates (in Table \ref{tab:template}, those prompts are task-related but not domain-relevant). Thus, the suboptimal performance of Perplection can also be explained by our hypothesis that generic templates are less effective at aligning the downstream data into a pre-trained feature space compared to those fine-grained domain-specific templates. We suspect that this can be addressed by providing Perplection with several domain-related fine-grained templates to select from. We leave these explorations for future work. All observations, however, show that it is effective to use perplexity to rate templates and select desired ones accordingly. 

\paragraph{Results on English datasets:} Table \ref{tab:eng_result} compares the performance of Perplection to random baselines on three English datasets. Perplection consistently tops the comparison in almost all cases except for SST-2 with RoBERTa. This observation supports the supposition that Perplection is agnostic to the pre-trained model used, and shows that it is promising to extrapolate results to other languages. \looseness=-1

\subsection{In-depth Analysis} 

\begin{figure}[!t]
\centering
\includegraphics[width=.49\textwidth]{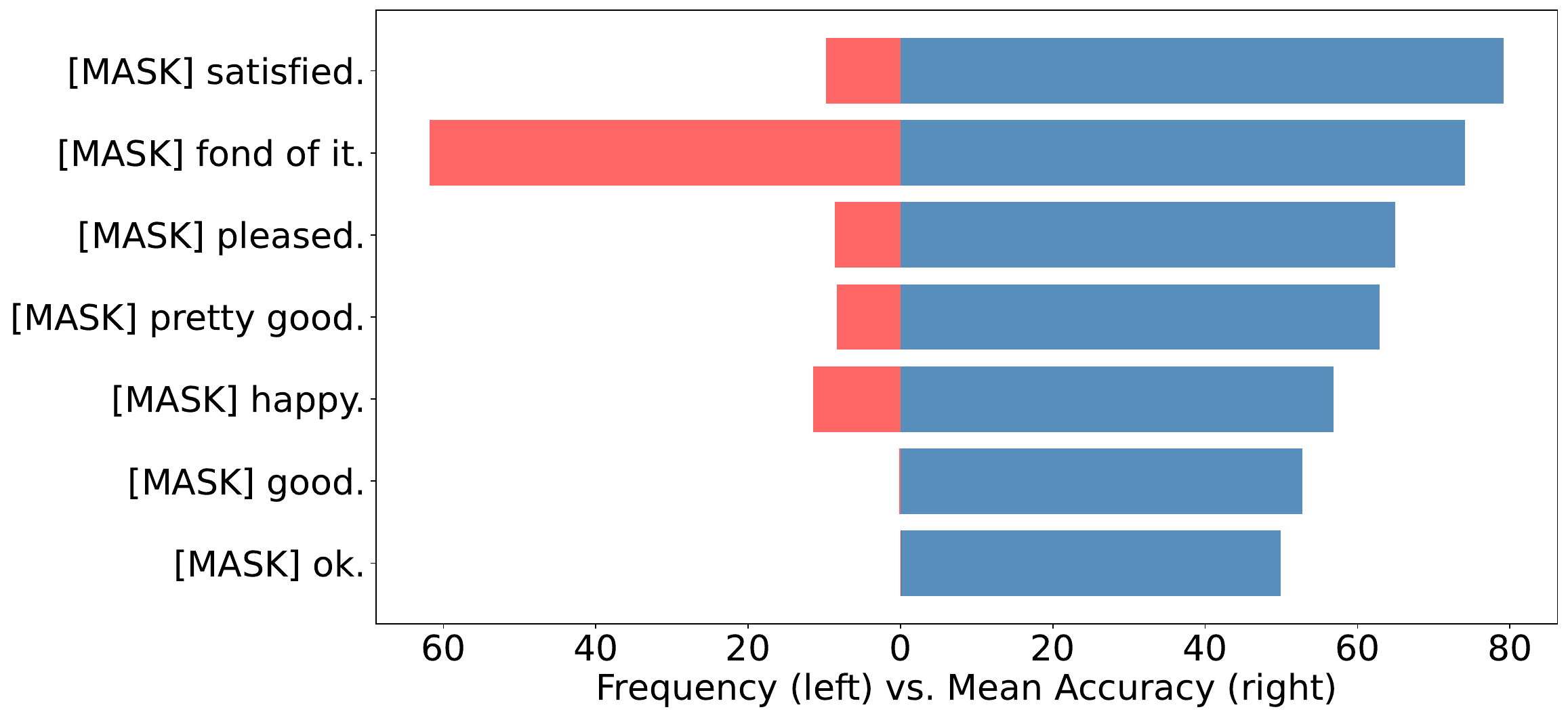}
 \caption{Normalised frequency of being selected vs. template quality measured by mean accuracy. }\label{fig:template_analysis}
\end{figure}

We conduct an in-depth analysis based on MPerplectionR. For brevity, we apply each manual prompting setting to all examples from the four datasets (i.e., \emph{DOUBAN}, \emph{WEIBO}, \emph{WAIMAI}, \emph{ECOMMERCE}) and aggregate the accuracy score as a post-hoc measurement of template quality. For each template, we also compute its frequency of being selected.
The results are presented in Figure \ref{fig:template_analysis}. It shows that templates with lower perplexity are more likely to achieve better performance. To be specific, there is 60\% chance for Perplection to select the second best performing template (i.e., \textit{``[MASK] fond of it.''}) and around 10\% chance to select the best performing template (i.e., \textit{``[MASK] satisfied.''}). For templates with no discriminative ability e.g., \textit{``[MASK] good.''} and \textit{``[MASK] ok.''}, our method has almost no chance to select them. Most importantly, the selection based on perplexity is annotation-agnostic and allows us to ``foresee'' the result to some extent without the need of a human-annotated development set. To conclude, the results demonstrate that perplexity is a reasonable metric for evaluating prompting settings.

\section{Discussion}

\paragraph{What contributes better zero-shot learners?}

This work empirically reveals that the large language discrepancy between the pre-training corpora and the downstream data may hinder the zero-shot generalization. On top of that, we develop a perplexity-based scheme that leverages cloze-style prompt templates to bridge language discrepancy and thus, fully releases the potential of pre-trained language models. The significance of this work lies in its pioneering study of a feasible objective for optimising REALISTIC zero-shot prompting templates. The idea may be applied to various variations (e.g., continuous prompts) beyond the discrete prompts currently being studied.

\paragraph{Why REALISTIC zero-shot matters?}

In this work, we constantly emphasise a realistic zero-shot scenarios (no labelled data), as opposed to the existing zero-shot setting in the field of NLP~\cite{xu2021fewclue,sun2021nsp} or Multi-modality~\cite{radford2021learning}, where a development set is available for template selection or hyper-parameter tuning. Realistic zero-shot can be quite appealing for industrial scenarios and thus, this research opens up a new avenue for research in the field of zero-shot learning, probably inspiring follow-up studies in broader tasks for advancing the zero-shot learning in industrial applications (especially in many low-resource scenarios). 

\paragraph{Potential impact in the LLM era.}

In light of the advancements in large language models (LLM) based on the decoder-only architecture \cite{zhao2023survey}, searching for effective instructions or in-context demonstration examples \cite{zhang-etal-2022-active} has become an essential challenge. Perplection can be seamlessly applied to decoder-only models for searching effective instructions/in-context examples for various natural language generation (NLG) tasks. We make our code available for replication and further extension to NLG tasks by the community.

% We emphasize that a realistc, which probably inspires follow-up studies in broader tasks for advancing the zero-shot learning in industrial applications (especially in many low-resource scenarios)

% The importance of this work lies in the fact that it is one of the first efforts to study a feasible objective for optimizing REALISTIC zero-shot prompting templates

% highlight the importance of this work spearheading the effort to study a feasible objective for optimizing  REALISTIC zero-shot prompting templates (may work with different variants beyond the discrete prompts here)

% 3. clarify realistic zero-shot setting (NLP/CV/VL?) vs exiting settings 

% Finally,  impact on the inference of huge language model, which probably inspires follow-up studies in broader tasks for advancing the zero-shot learning in industrial applications (especially in many low-resource scenarios). 

\section{Conclusion}

We developed Perplexity Selection Prompt (Perplection) a method that enables real-world zero-shot text classification without the use of any human-annotated data. A pilot study demonstrated that Perplexity can be an effective measure of the efficacy of templates. Experimental results show that, for datasets in both English and Chinese, our method can boost zero-shot performance of cloze-style prompt learning in binary sentiment analysis as well as multi-class classification, without using a development set. Further in-depth analysis supports the observation  that Perplection can ``foresee'' the efficacy of prompt templates. 

% This work present a solid step towards investigating the underlying rationales behind cloze-style prompt learning in zero-shot scenarios. To this end, we elaborate why fine-tuning is suboptimal in few-shot and propose a theoretical framework to explain the efficacy of prompt learning. Comprehensive experiments are performed to support proposed assumptions. Finally, as an extension of the theoretical framework, we propose using perplexity to measure the quality of prompting setting. Experimental results show that this method leads to significant prediction benefits and allows us to ``forecast'' the ability of model prior to inference.

\section{Limitations} 

In this study, we mainly utilised the BERT family of models for Chinese text classification tasks. Given the similarity with respect to transformer language models and pre-training paradigms, as well as the preliminary results on English datasets as discussed in Section \ref{subsect:exp_results}, we may be able to extrapolate the results to other architectures/tasks/languages. 

For example, Perplection can be seamlessly apply to decoder-only models (e.g., GLM \cite{du-etal-2022-glm}, LLaMA \cite{touvron2023llama}) to see whether it can boost the performance for those NLG tasks. But further investigation is needed to verify the utility of findings on other model architectures, tasks, and languages. In the future, we expect to see Perplection applied to different NLG tasks such as seq2seq information extraction \cite{Lu2022PUnifiedNERAP}, question answering, arithmetic reasoning, machine translation or even multi-modality tasks.

Also, utilising Perplection may exacerbate the inherent limitations of pre-trained language models. We suspect that, in instances where the model has not been exposed to certain texts or concepts during pre-training, reliance on perplexity for template selection may result in subpar performance. In the future, we expect to explore whether we can alleviate this problem by certain annotation-free methods, such as continuous self-supervised training with downstream data, or extend our method in a few-shot setting where limited label information is available. 

Besides, the use of perplexity as a metric has the drawback of favoring long texts, which forces us to design templates with the same length. Therefore, a length-agnostic metric can be considered as an alternative.

% 1. magnify the bias in pretrained model
% 2. automation generated template is bad

\section{Ethics Statement}
We honor the ACL Code of Ethics. No private data or non-public information was used in this work. We conducted our research in an objective and unbiased manner. We take full responsibility for the content of this paper and stand behind the accuracy and integrity of our work.

\section*{Acknowledgements}
We would like to thank anonymous reviewers for their insightful comments to help improve the paper. This publication has emanated from research conducted with the support of SenseTime Research.

\bibliography{custom}
\bibliographystyle{acl_natbib}

\appendix

\begin{table*}[!htb]
\resizebox{1.0\textwidth}{!}{%
\centering
\begin{tabular}{lllll}
\toprule
 & \textbf{Text in Chinese}                                         & \textbf{Translation}                                                                                                                                             & \textbf{Perplexity} &  \\ \midrule

 & 阿姨：不要太累了{[}哈哈{]}                                                                     & Auntie: Don't be too tired {[}haha{]}                                                                                                                                                & 17.21                                   &  \\ \midrule
 & \begin{tabular}[c]{@{}l@{}}撒娇大法，啊的身份拉升大盘撒娇大法，啊\\ 的身份拉盘。阿姨：不要太累了{[}哈哈{]}\end{tabular} & \begin{tabular}[c]{@{}l@{}}Coquetry Dafa, ah's identity pulls up the big market Coquettish Dafa,\\ ah's identity pulls the plate. Auntie: Don't be too tired {[}haha{]}\end{tabular} & 5.85                                    &  \\\bottomrule
 &                                                                                      &                                                                                                                                                                                      &                                         & 
\end{tabular}%
}
\caption{Comparison of a long nonsense sentence with a short fluent sentence.}
\label{tab:ppl_drawback_comp}
\end{table*}

% Please add the following required packages to your document preamble:
% \usepackage{graphicx}
\begin{table*}[!htb]
\resizebox{\textwidth}{!}{%
\begin{tabular}{ll}
\toprule
\textbf{Dataset} & \textbf{Mapping}                                                                                                                                                                                                                                              \\ \midrule
TNEWS            & \begin{tabular}[c]{@{}l@{}}\{100:'故事' (story),101:'文化' (cultural),102:'娱乐' (entertainment),103:'体育' (sports),\\104:'财经' (finance),106:'房产' (real estate),107:'汽车' (automobile),108:'教育' (education),\\ 109:'科技' (technology),110:'军事' (military),112:'旅游' (trip),113:'国际' (world-wide),\\114:'股票' (stock),115:'农业' (agricultural),116:'电竞' (e-sports)\}\end{tabular}                                                                         \\ \midrule
CSLDCP           & \begin{tabular}[c]{@{}l@{}}\{'材料科学与工程': '材料' (Materials),'力学': '力学' (Mechanics),\\'园艺学': '园艺' (Horticulture),'水产': '水产' (Aquaculture), '航空宇航科学与技术': '航空' (Aerospace Science),\\ '建筑学': '建筑' (Architecture),'林学/林业工程': '林业' (Forestry ), '天文学': '天文' (Astronomy), \\'机械工程': '机械' (Mechanical),'地理学': '地理' (Geography), '大气科学': '大气' (Atmospheric Science),\\'测绘科学与技术': '测绘' (Geodesy),'军事学': '军事' (Military Science),'新闻传播学': '新闻' (Journalism),\\'植物保护': '植物' (Plant)\}\end{tabular} \\ \midrule
IFLYTEK          & \begin{tabular}[c]{@{}l@{}}\{107: '团购' (group buy),110: '超市' (supermarket),113: '办公' (office),18: '动作' (motion),2: '免费' (free),\\30: '情侣' (dating),3: '租车' (ride-hailing),42: '百科' (encyclopedia),48: '音乐' (music), 64: '民航' (airline),\\75: '汽车' (automobile), 87: '美妆' (makeup),89: '餐饮' (food),91: '运动' (fitness),92: '支付' (payment)\}\end{tabular}                                                                      \\ \bottomrule
\end{tabular}%
}

\caption{The mapping of class names to label words with equal length. Translations are provided in brackets.}
\label{tab:template_mapping_words}
\end{table*}

\begin{table*}[!htb]
\resizebox{\textwidth}{!}{%
\begin{tabular}{l|l|l|l}
\toprule
\textbf{Task}                                                                                                                   & \textbf{Perplection}                                                                                                                                                                                                                                                                                                                                                                                                                                                                                                                                                                                                                                                                                                                                                                                                                                                                                                        & \textbf{Zero-PET}                                                                                                                                                        & \textbf{NSP-BERT}                                                                                                                                                         \\ \midrule
\begin{tabular}[c]{@{}l@{}}Sentiment Analysis datasets \\ (i.e., WAIMAI, WEIBO,  \\ DOUBAN, ECOMMERCE, \\ EPRSTMT)\end{tabular} & \begin{tabular}[c]{@{}l@{}}Template1: {[}MASK{]}满意。 ({[}MASK{]} satisfied.)\\ Template2: {[}MASK{]}喜欢。 ({[}MASK{]} font of it.)\\ Template3: {[}MASK{]}高兴。 ({[}MASK{]} pleased.)\\ Template4: {[}MASK{]}可以。 ({[}MASK{]} pretty good.)\\ Template5: {[}MASK{]}开心。 ({[}MASK{]} happy.)\\ Template6: {[}MASK{]}好。     ({[}MASK{]} good.)\\ Template7: {[}MASK{]}行。     ({[}MASK{]} ok.)\\ \\ Label words: 很;不 (very; not)\end{tabular}                                                                                                                                                                                                                                                                                                                                                                                                                                                                                         & \begin{tabular}[c]{@{}l@{}}Template: 这次买的东西很{[}MASK{]}。 \\ (The things I bought this time is very {[}MASK{]}.)\\ \\ Label words: 好;差 (good; bad)\end{tabular}            & \begin{tabular}[c]{@{}l@{}}Template: 这次买的东西很{[}MASK{]}.  \\ (The things I bought this time is very {[}MASK{]}.)\\ \\ Label words: 好;差 (good; bad)\end{tabular}            \\ \midrule
TNEWS                                                                                                                           & \multirow{3}{*}{\begin{tabular}[c]{@{}l@{}}Template1: 这属于是{[}MASK{]}。  (This belongs to {[}MASK{]})\\ Template2: 此话属于{[}MASK{]}。  (The words belong to {[}MASK{]})\\ Template3: 实际上，{[}MASK{]}。  (Actually it is {[}MASK{]})\\ Template4: 应该算是{[}MASK{]}。  (Probably it is {[}MASK{]})\\ Template5: 方向为{[}MASK{]}。      (The direction is {[}MASK{]})\\ Template6: 归功于{[}MASK{]}。      (This is due to {[}MASK{]})\\ Template7: 给它放到{[}MASK{]}。  (Put it into {[}MASK{]})\\ Template8: 它意思是{[}MASK{]}。  (It means {[}MASK{]})\\ Template9: 明显算{[}MASK{]}。      (Obviously counted as {[}MASK{]})\\ Template10: 显而易见{[}MASK{]}。(Obviously it is {[}MASK{]})\\ \\ Label words (TNEWS): 故事;文化;娱乐 ...  \\ (story; cultural; entertainment ...)\\ \\ Label words (CSLDCP): 材料;力学;园艺 ... \\ (Materials; Mechanics; Horticulture...)\\ \\ Label words (IFLYTEK): 团购;超市;办公 ... \\ (group buy; supermarket; office...)\end{tabular}} & \begin{tabular}[c]{@{}l@{}}\\ \\ \\Template: 这是一则{[}MASK{]}新闻。 (This is a {[}MASK{]} news.)\\ \\ Label words: 故事;文化;娱乐 ... (story; cultural; entertainment...)\\ \\ \\\end{tabular}      & \begin{tabular}[c]{@{}l@{}}\\ \\ \\Template: 这是一则{[}MASK{]}新闻.  (This is a {[}MASK{]} news.)\\ \\ Label words: 故事;文化;娱乐 ... (story; cultural; entertainment...)\\ \\ \\\end{tabular}      \\ \cline{1-1} \cline{3-4} 
CSLDCP                                                                                                                          &                                                                                                                                                                                                                                                                                                                                                                                                                                                                                                                                                                                                                                                                                                                                                                                                                                                                                                                             & \begin{tabular}[c]{@{}l@{}}\\ \\ \\Template: 这是一篇{[}MASK{]}论文。 (This is a {[}MASK{]} paper.)\\ \\ Label words: 材料;力学;园艺 ... (Materials; Mechanics; Horticulture...)\\ \\ \\\end{tabular} & \begin{tabular}[c]{@{}l@{}}\\ \\ \\Template: 这是一则{[}MASK{]}论文.  (This is a {[}MASK{]} paper.)\\ \\ Label words: 材料;力学;园艺 ... (Materials; Mechanics; Horticulture...)\\ \\ \\\end{tabular} \\ \cline{1-1} \cline{3-4} 
IFLYTEK                                                                                                                         &                                                                                                                                                                                                                                                                                                                                                                                                                                                                                                                                                                                                                                                                                                                                                                                                                                                                                                                             & \begin{tabular}[c]{@{}l@{}}\\ \\ \\Template: 这是一款{[}MASK{]}类软件。(This is a {[}MASK{]} app.)\\ \\ Label words: 团购;超市;办公 ... (group buy; supermarket; office...)\\ \\ \\\end{tabular}       & \begin{tabular}[c]{@{}l@{}}\\ \\ \\Template: 这是一则{[}MASK{]}类软件.  (This is a {[}MASK{]} app.)\\ \\  Label words: 团购;超市;办公 ... (group buy; supermarket; office...)\\ \\ \\\end{tabular}      \\ \bottomrule
\end{tabular}%
}
\caption{Manually generated templates and label words for Perplection, and other baselines Zero-PET and NSP-BERT. For Perplection and Zero-PET, we prefix the template. For NSP-BERT, we suffix the template as suggested in~\cite{sun2021nsp}. Due to space considerations, we have omitted some label words, which can be referred to in Table \ref{tab:template_mapping_words}. Translations are provided in brackets.}
\label{tab:baseline_setting}
\end{table*}

\begin{table*}[!htb]
\centering
\resizebox{.7\textwidth}{!}{%
\begin{tabular}{l|l|l}
\toprule
\textbf{Dataset} & \textbf{Templates}                                                                                                                                                                                                                                                                                                                                                                                                       & \textbf{Label Words}                                 \\ \midrule
SST-2            & \begin{tabular}[c]{@{}l@{}}Template1: that sounds like {[}MASK{]}\\ Template2: this is obviously {[}MASK{]}\\ Template3: it should be {[}MASK{]}\\ Template4: actually, it's {[}MASK{]}\\ Template4: in fact, it's {[}MASK{]}\\ Template5: it's very {[}MASK{]}\\ Template6: it is {[}MASK{]}\\ Template7: I mean it's {[}MASK{]}\\ Template8: it means {[}MASK{]}\\ Template10: I think {[}MASK{]}\end{tabular}         & \{'negative': 'negative', 'positive': 'positive'\}   \\ \midrule
TweetEval        & \begin{tabular}[c]{@{}l@{}}Template1: that sounds like {[}MASK{]}\\ Template2: this is obviously {[}MASK{]}\\ Template3: it should be {[}MASK{]}\\ Template4: actually, it's {[}MASK{]}\\ Template4: in fact, it's {[}MASK{]}\\ Template5: it's very {[}MASK{]}\\ Template6: it is {[}MASK{]}\\ Template7: I mean it's {[}MASK{]}\\ Template8: it's like {[}MASK{]}\\ Template10: whatever it is {[}MASK{]}\end{tabular} & \{0: 'positive', 1: 'negative'\}                     \\ \midrule
AG News          & \begin{tabular}[c]{@{}l@{}}Template1: this is {[}MASK{]}\\ Template2: it is {[}MASK{]}\\ Template3: I mean {[}MASK{]}\\ Template4: actually, answer is {[}MASK{]}\\ Template5: it should be {[}MASK{]}\\ Template6: in fact, it's {[}MASK{]}\\ Template7: the sentence is {[}MASK{]}\\ Template8: it belongs to {[}MASK{]}\\ Template9: this news is {[}MASK{]}\\ Template10: in my opinion {[}MASK{]}\end{tabular}      & 0: 'world', 1: 'sports', 2: 'business', 3: 'science' \\ \bottomrule
\end{tabular}%
}
\caption{Manually generated templates and label words for Perplection in English datasets.}
\label{tab:eng_template}
\end{table*}

\section{Issue of Perplexity}\label{appendix:perplexity}

\begin{figure}[!htb]
\centering
\includegraphics[width=.45\textwidth]{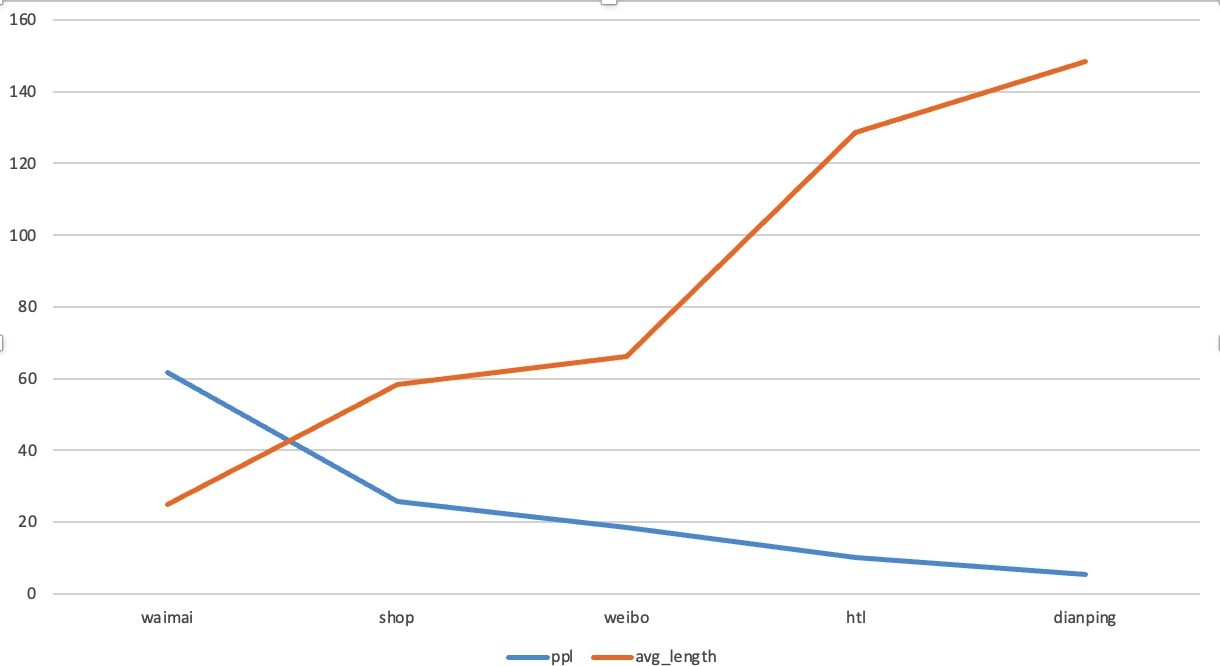}
 \caption{Line chart of average perplexity and average text length across different datasets. The x-axis represents the dataset, the blue line is the mean perplexity score while the orange line is the mean text length.}\label{fig:ppl_textlength}
\end{figure}

We find that the current perplexity definition has the drawback of favouring longer sentences. That is, a sentence is assigned a lower perplexity, not because the pre-trained language model can more easily model this sentence (i.e., lower language discrepancy), but rather because the text is longer. We first use a simple comparison to demonstrate this as shown in Table \ref{tab:ppl_drawback_comp}. We calculate the perplexity of a meaningful sentence \textit{``Auntie: Don't be too tired [haha]''} which is 17.21. However, if we prefix this sentence with a long sequence of nonsense words, the perplexity even gets lower, i.e., 5.85. We then conduct a large scale test to see the correlation between perplexity and text length. The results are presented in Figure \ref{fig:ppl_textlength}, it is obvious that the avg. perplexity is inversely proportional to avg. text length. In other words, a low perplexity of a sentence is partially contributed by a low language discrepancy but more likely to be contributed by a long text, which challenges our use of perplexity to measure language discrepency.

\section{Automatic Template Generation}\label{appendix:templates}

Similar to~\citet{gao-etal-2021-making}, for the  \emph{DOUBAN}, \emph{WEIBO}, \emph{WAIMAI}, and \emph{ECOMMERCE} datasets we fix the verbaliser to \{\textit{very}: $++$, \textit{not}: $--$\}, and use T5-v1.1-base-chinese\footnote{\url{https://huggingface.co/uer/t5-base-chinese-cluecorpussmall}.} to automatically generate templates. Specifically,~\citet{gao-etal-2021-making} assume a few-shot scenario using ground truth label word as well as corresponding examples to generate a number templates. They then sort generated templates based on the aggregated generation probability (the calculation of generation probability also needs label information) of the whole training set. However, our experiment assumes a zero-shot scenario with no labelled data. Thus, for each dataset, we first randomly sample 50 examples from the pool. For each example, we use label words indicating both sentiments to generate templates, one for each sentiment, resulting in 100 templates in total. Then we remove duplicate templates, leaving around 59-73 templates remain per dataset respectively. 

For the \emph{EPRSTMT}, \emph{TNEWS}, \emph{CSLDCP}, and \emph{IFLYTEK} datasets, whose automatically generated templates have been made available,\footnote{\url{https://github.com/CLUEbenchmark/FewCLUE/tree/main/baselines/models_pytorch/LM-BFF/my_auto_template.}}, we directly use those existing generated templates. We remove duplicate templates and around 11-22 templates remain per dataset. All automatically generated templates can be seen at URL masked for anonymous review.

\begin{table}[!htb]
\small
\centering
\resizebox{.5\textwidth}{!}{%
\begin{tabular}{lcccccccc}
\toprule
\multirow{2}{*}[-3pt]{\textbf{Datasets}} & \multicolumn{3}{l}{\textbf{1. {[}very/not{]} pleased.}} & \textbf{} & \multicolumn{3}{l}{\textbf{2. {[}yellow/red{]} black.}}  \\[3pt] \cline{2-4} \cline{6-8} \\[-5pt]
                                  & \textbf{PPL\textsubscript{g}}         & \textbf{PPL\textsubscript{r}}         & \textbf{Diff.} & & \textbf{PPL\textsubscript{g}}        & \textbf{PPL\textsubscript{r}}       & \textbf{Diff.}            \\ \midrule
\textit{Douban}                    & 24.10                    & 25.12                                &   -1.02   &     & 67.91                    & 74.11  & -6.20                      \\
\textit{Weibo}                     & 19.17                   & 20.39                                 &  -1.22     &     & 44.39                    & 44.51   & -0.12                       \\
\textit{Waimai}                    & 16.06                   & 16.82                         &   -0.76   &     & 22.60                  & 24.07     & -0.20                                 &                                  \\
\textit{Online-shopping}           & 13.55                   & 14.58                               &   -1.03   &     & 28.51                    & 28.61   & -0.10                       \\ \bottomrule
\end{tabular}%
}
\caption{Mean perplexity of prompting with ground truth label word (PPL\textsubscript{g}), prompting with reversed label word (PPL\textsubscript{r}), and difference between two templates computed by PPL\textsubscript{g} minus PPL\textsubscript{r} (Diff.). }
\label{tab:reverse}
\end{table}

\section{Implementation Details}\label{appendix:implemenation}

In the implementation of Zero-PET, we use the pre-trained Chinese-RoBERTa-wwm-ext model, which is identical to the model employed in Perplection. For NSP-BERT, we use google BERT-Chinese.\footnote{\url{https://huggingface.co/bert-base-chinese.}} Templates and label words for both baselines follow the best-performing setting reported in~\cite{sun2021nsp,xu2021fewclue}, as shown in Table \ref{tab:baseline_setting}. The manual generated templates (in Chinese) for Perplection are also shown in Table \ref{tab:baseline_setting}. A conversion is conducted to map class names to label words following~\cite{xu2021fewclue} to ensure all prefixed texts have similar length, as shown in Table \ref{tab:template_mapping_words}. For the \emph{CSLDCP} and \emph{IFLYTEK} datasets we randomly subsample 15 classes to facilitate the experiments.

In the implementation of English Perplection and its random counterparts, we use the pre-trained BERT-base-uncased\footnote{\url{https://huggingface.co/bert-base-uncased}} and RoBERTa-base\footnote{\url{https://huggingface.co/roberta-base}} models. Templates and label words for English experiments are shown in Table \ref{tab:eng_template}.
All experiments are conducted on a Tesla V100 GPU with 32GB memory.

\section{Reverse Label Words}\label{appendix:reverse_exp}

To briefly verify whether perplexity can be used to measure the quality of prompting, we perform a very simple experiment where we compute the mean perplexity score of prompted input $x'$ with \textit{``[MASK]''} filled by ground truth label words for each dataset (called \textbf{PPL\textsubscript{g}} ). Then we reverse the label words filled in previous input examples (e.g., we change \textit{``very pleased.''} to \textit{``not pleased.''} in a positive sentiment example) and recompute mean perplexity score (called \textbf{PPL\textsubscript{r}}). Note that this experiment is based on RoBERTa. The results of this are shown in Table \ref{tab:reverse}. 

First, we notice that in Setting 1 (i.e., \textit{``[very/not] pleased.''}), the mean perplexity of PPL\textsubscript{g} is always smaller than that of PPL\textsubscript{r} by a clear margin which is encouraging. This shows that the pre-trained model can perceive the change of semantics in texts. When we see the perplexity of Setting 2 (i.e., \textit{``[yellow/red] black.''}, we find out the magnitude of change is much smaller, which demonstrates that replacing label words makes almost no difference to models if domain-irrelevant prompting is applied.

\end{CJK}
\end{document}